# Computation of Diet Composition for Patients Suffering from Kidney and Urinary Tract Diseases with the Fuzzy Genetic System

Sri Hartati

Department of Electronics and Computer Sciences, Faculty of Math and Natural Sciences, Gadjah Mada University, Sekip Utara, Bulaksumur, Yogyakarta, Indonesia

Shofwatul 'Uyun

Program of Informatics, Faculty of Sciences and Technology, Sunan Kalijaga State Islamic University, Marsda Adisucipto Street No.1 Yogyakarta, Indonesia

## ABSTRACT
Determination of dietary food consumed a day for patients with diseases in general, greatly affect the health of the body and the healing process, is no exception for people with kidney disease and urinary tract. This paper presents the determination of diet composition in the form of food subtance for people with kidney and urinary tract diseases with a genetic fuzzy approach. This approach combines fuzzy logic and genetic algorithms, which utilizing fuzzy logic fuzzy tools and techniques to model the components of the genetic algorithm and adapting genetic algorithm control parameters, with the aim of improving system performance. The Mamdani fuzzy inference model and fuzzy rules based on population parameters and generation are used to determine the probability of crossover and mutation, and was using In this study, 400 food survey data along with their substances was used as test material. From the data, a varying amount of population is established. Each chromosome has 10 genes in which the value of each gene indicates the index number of foodstuffs in the database. The fuzzy genetic approach produces 10 best food substance and their compositions. The composition of these foods has nutritional value in accordance with the number of calories needed by people with kidney and urinary tract diseases by type of food.

## 1. INTRODUCTION
To achieve and maintain optimal health and nutritional status, humans body needs to consume daily diet containing balanced nutrients. When the body can digest, absorb and metabolize nutrients well, it will reach a state of balanced nutrition. But in ill condition, dietary modifications pursued in order to remain balanced nutrition. Balanced Nutrition General Guidelines (PUGS) is a basic guide balanced nutrition prepared as a guide on food consumption behavior in the community as a absobaik and correct. *Pedoman Umum Gizi Seimbang* (PUGS) or *General Guidance for Balanced Nutrient* is a handbook to guide people for having balanced nutrition in the diets. In the Guidance, food ingredients are classified based on three functions; as sources of energy or power, as sources of protein, and as sources of stabilizers (such as vegetables and fruits). PUGS suggests that a good food composition comprises of 60%-75% carbohydrate, 10%-15% protein and 10%-25% fat. The term of nutrition value refers to the amount of food composition required to maintain the level of adequate nutrition. People need to consider their nutritional needs based on age, gender, and physical activities. Special conditions apply in determining nutritional composition. In this regard, people with deteriorating health conditions due to infection, metabolism disturbance, chronic diseases, and other abnormalities need a special calculation before determining their diets. In the case of patients with kidney and urinary tract diseases would need to carefully measure their nutritional intake to meet the nutritional balance [1]. The measurement and determination can be calculated by using computerized system of soft computing techniques.

Various calculations can be applied simultaneously to improve the effectiveness of the method, for example using the *neural network* and *genetic algorithm* at a system to solve a particular problem. The *Backpropagation* method, one of learning methods in the *neural network*, can be used to improve convergent score in genetic algorithm for searching a global and optimal solution. According to Javadi et.al [2] integrating *neural network* with *genetic algorithm* may significantly accelerate the convergence of genetic algorithm and can certainly improve the quality of the result.

The best practice of using genetic algorithm for determining the alphabet of amino acids has been done by Palensky et.al. [3]. Simplifying the alphabet of amino acids has been successfully conducted in several fields of *bioinformatics*, including: the prediction of protein structure, the prediction of protein function and the prediction of protein classification. Another benefit of using genetic algorithm is to deal with a big set of data. In this case, genetic algorithm may offer several best solutions. However, genetic algorithm has a weakness in itself. It uses trial and error in search of the best solution.

Genetic algorithm which is capable in generating optimal output can overcome its own weakness by using an optimizing concept for improving the effectiveness of the system. One of the ways is to combine the fuzzy logic method and the genetic algorithm. Some research have proved that there is an improvement of the system by using the genetic fuzzy [4,5,6].

In this research, the genetic fuzzy is developed as a model to determine the composition of food ingredients, as a dietary guidance for those suffering from kidney and urinary tract diseases. Thus, it can be easier for nurses to take care of their patients. The aim of such a diet is to improve the quality of the meals and sport activities, in order to have a better metabolic control of the patients.

## 2. FUZZY GENETIC
The fuzzy system developed here is a decision making modelled that takes a fuzzy and a crisp input. Input from a crisp number will be transformed into a fuzzy number. This process is called *fuzzification*. Then, it is processed using fuzzy rules to produce a fuzzy number as an output. The process to generate an output from the fuzzy system resulting on a crisp number is called *defuzzification*. Mamdani's model was introduced firstly by Ebrahim Mamdani in 1975, which is well known as the Max-Min method; using Min for implication functions and using Max for composing between implication functions. The computing process is so complex and therefore needs to spend much time, which makes the result highly accurate [7].

The Genetic Algorithm (GA) is firstly proposed by John Holland and his colleagues at the University of Michigan for an automatic cellular application. The GA application is usually applied to solve the problem of a complex optimizing task, for instance at a shop scheduling, an image processing and a combinatorial optimizing, an application that requires an adaptable problem solving strategy (Gen et.al.) [8]. Generally speaking, the genetic algorithm has 5 basic components, as stated by Michalewicz [9]:
 (a) A genetic representation of various solutions from a problem
 (b) A way to generate population initialization from various solutions
 (c) An evaluation of a solution function with a fitness score that belongs to each individual
 (d) Several genetic operators that divide several off springs during a reproduction process
 (e) A score for several parameters from the genetic algorithm

The genetic algorithm is a computational model inspired by evolution process. It can be applied in various complex fields, such as in engineering designs and modern operational system and a system as proposed by Ferentinos and Tsiligiridis [10]. In addition, the genetic algorithm is a technique to look for a stochastic based on population and optimizing algorithm which adopts the evolution paradigms. The genetic algorithm does natural selections and uses several genetic operators, such as crossover, mutation etc. A pseudo-code from the genetic algorithm is shown in the picture 1.

  Define the generation of t ← 0
  Define the initial population P (0) randomly
  Evaluate all individuals in P (0)
  **Replicate**
    Choose some individuals of P (t) which are potential for a crossover
    Do the crossover to produce off springs
    Do the mutation process for the off springs
    Change the P (t) with a new population
    Define it as generation t ← t+1
    Evaluate all individuals in P (t)
  **Until** all criteria are met

**Fig 1 :. A Pseudo-code from the genetic algorithm**

The genetic fuzzy is an integration between a logic fuzzy and a genetic algorithm. There are two types of genetic fuzzy. The first type is involved in the genetic algorithm for solving the problem of optimizing and searching related to the fuzzy system. The second type is using the fuzzy tool and the logical technique of fuzzy to model the genetic algorithm components and to adapt determinant parameters of genetic algorithm that aim at improving the system. The current research will deal with the second type.

## 3. THE NUTRIENT NEEDS FOR THOSE SUFFERING FROM KIDNEY AND URINARY TRACT DISEASES
Nutrient is an organic substance needed by an organism to function normally, to grow, and to keep healthy. Nutrition can be taken from food and drink, which will be then absorbed by the body. Researches in the field of nutrition deal with the impacts of food and beverages to the body, especially in determining the optimum diets. A thoughtful choice of daily diets will provide all the nutrients needed for the body to function normally. On the contrary, if the food is not well chosen, people will suffer from a deficiency of essential nutrients [11]. Nutritional needs are influenced by age, gender, physical activities, and special conditions such as pregnancy and lactation. There are several ways to determine the number of Basal Metabolism (NBM), one of them is the postulate of Harris Benedict [1] that will be applied in this research.

   Males   = 66+(13,7xBW)+(5xH)-(6,8xA)      (1)
   Females = 655+(9,6xBW)+(1,8xH)–(4,7xA)    (2)
Explanation :
   BW = body weight (kg)
   H  = height (cm)
   A  = Age
The formula (3) can be used to calculate the amount of energy.

Energy needs = AMB x activity factors x stress factor   (3)

There are two activity factors for those who are sick; bed rest, with 1, 2 factor; and activities not related with the bed, with 1, 3 factor. Meanwhile, there are six types of trauma; (1) condition without stress, the patient is in a good condition, the score of the factor is 1,3; (2) minor stress: gastrointestinal ulcers, cancer, elective surgery, moderate skeletal injuries, the score of the factor is 1,4; (3) medium level stress : sepsis, bone surgery, burn injuries, major skeletal injuries, the score is 1,5; (4) *major stress*: multiple trauma, sepsis and multisystem surgery, the score is 1,6; (5) *very serious stress:* serious head injuries, acute respiratory diseases, burn injuries and sepsis, the score is 1,7; and lastly (6*) very serious burn injuries, the score is* 2,1.

The main function of kidney is to maintain the balance of homeostatic liquid, ions and organic substances. A special diet is required if the function of the kidney is deteriorated due to the following diseases: (1) nefrotic syndrome (2) acute kidney failure (3) chronic kidney diseases with de-functioning of the kidney, from minor to major degrees of illness; (4) the final phase of kidney diseases requiring kidney transplantation and (5) kidney stone. The diet of a kidney disease is emphasized on controlling the consumption of energy, protein, liquid, ions such as sodium, potassium, calcium and phosphorous.

The requirement of diet for nefrotic syndrome with mild edema is as follows: (1) energy = 35 kcal/kg BW per day; (2) protein= 1, 0 g/kg BW; (3) fat = 15-20% from the amount of required total energy; (4) Sodium = 1 g per day; (5) cholesterol < 300 mg. The diet for acute kidney disease with mild *catabolic* and without *anuria* is: (1) energy = 30 kcal/kg BW; (2) protein = 0,8 g/kg BW; (3) fat 25% of the total required energy. The diet for chronic kidney with *hyperkalemia* is: (1) energy = 35 kcal/kg BW; (2) protein = 0,7 g/kg BW; (3) fat = 25 % of the total required energy; (4) potassium = 55 mEq. The diet for the kidney transplantation in the first month after the transplantation process is: (1) energy = 32,5 kcal/kg BW/day; (2) protein = 1,4 g/kg BW/day; (3) fat = <30 % of the total required energy; (4) calcium = 1000 mg/day; (5) phosphorous = 1000 mg/day. Meanwhile, the requirement for renal failure with dialysis is (1) energy = 35 kcal/kg BW; (2) protein = 1,1 g/kg BW ideal/day; (3) carbohydrate = 65 % of the total required energy; (4) fat = 22,5 % of the total required energy; (5) calcium = 1000 mg/day and phosphorous < 17 mg/kg BW ideal/day [1].

## 4. THE PROPOSED MODEL
### 4.1 General Explanation of the Model

The determination of crossover and mutation probability for choosing the diet composition of patients with kidney and urinary tract diseases uses a fuzzy model of Mamdani's inferential system. The parameters of genetic algorithm (GA) should be carefully defined to achieve a convergent result at a quick global optimum. Four parameters are sensitive to the performance of GA, i.e the scale of the population, *selective pressure* of the parental selection, crossover probability and mutation probability [12]. The fuzzy model of Mamdani's inferential system applied here consists of four variables; input and output variables. The input variable is population and generation measurement, and the output variable is crossover dan mutation probability. The components of the inferential system can be shown in the following picture

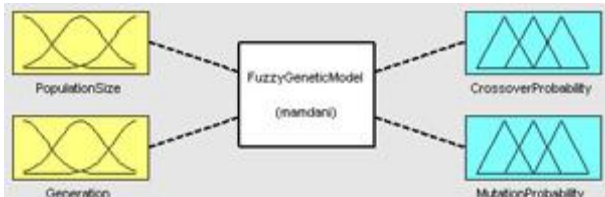

**Fig 2 : Mamdani's method inferential system**

Input variable for the measurement of the population has the universe context of speaking [0-150] that is divided into three fuzzy groups; small, medium and big. The function applied is a sigmoid curve and a phi and can be shown in the following picture:

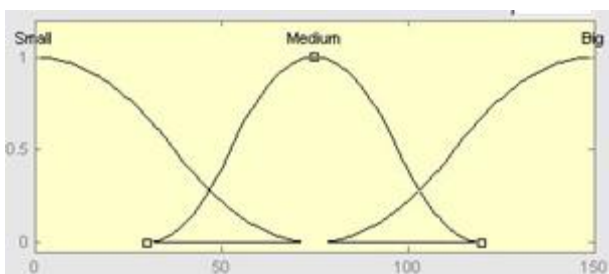

**Fig 3 : A function chart for population sized-variable**

The input variable for the generation has the universe context of speaking [0-1500] which is divided into three fuzzy groups; short, medium and long. The function applied is a sigmoid and a phi curves, as shown by the following picture.

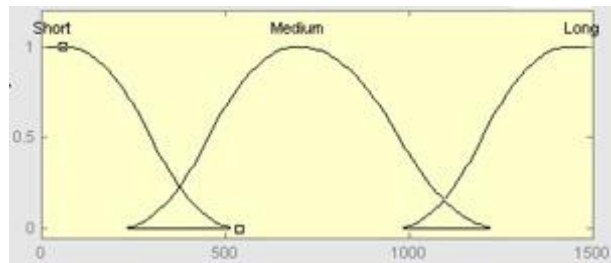

**Fig 4 : The Membership Function of the Variables of the Generation**

The output variable for crossover probability has universe context of speaking [0-1] which is divided into four fuzzy groups; small, medium big and very big. The function applied is sigmoid and phi curves, as shown by the following picture.

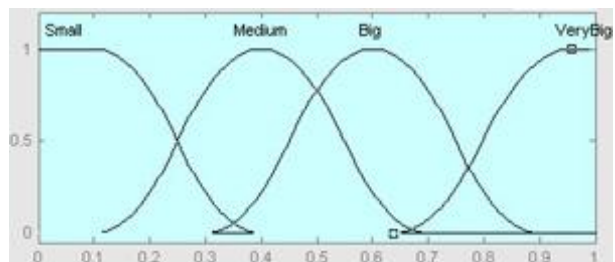

**Fig 5 : Graph of the membership function for the crossover probability variable**

The output variable for mutation probability has universe context of speaking [0 1] which is divided into four fuzzy groups; very small, small, medium, and big. The membership function applied is sigmoid and phi curves, as shown by the following picture.

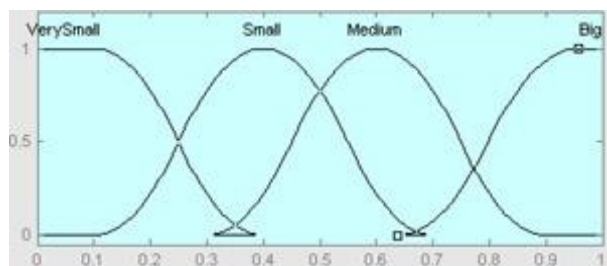

**Fig 6 : Graph of membership function for mutation probability variable**

The mechanism of the Mamdani's model of fuzzy inferential system is to receive an input in the form of a crisp number in which the process of *fuzzification* is then conducted based on 18 rules of aggregation process. The aggregation process for Mamdani's model uses MIN (intersection operator) at the function of implication and MAX (union operator) at the composition between the function of implication. The output of the aggregation process in the form of fuzzy number goes through a *defuzzification* process to produce a crisp number output.

The output of the fuzzy inferential system is then used for genetic algorithm to get the most optimum fitness function resulted in the composition of food ingredients that should be consumed by the patients with kidney and urinary tract diseases in a day. There are two ways of developing fuzzy rules, by using expert knowledge or automatic learning by benefiting from meta-genetic algorithm. This research uses the approach proposed by Xu and Vukovich [13] shown in the two following tables.

**Table 1 Fuzzy Decision of Crossover Probability (Cp)**

| Crossover Probability | Population size | | |
|---|---|---|---|
| Generation | Small | Medium | Big |
| Short | Medium | Small | Small |
| Medium | Big | Big | Medium |
| Long | Very big | Very big | Big |

Based on the above table, it can be found 9 fuzzy rules, as follows:
```
IF the generation is short AND the
   population size is small THEN the
   crossover probability is medium
IF the generation is medium AND the
   population size is small THEN the
   crossover probability is big
IF the generation is long AND the population
   size is small THEN the crossover
   probability is very big
IF the generation is short AND the
   population level is medium THEN the
   crossover probability is small
IF the generation is medium AND the
   population size is medium THEN crossover
   probability Pc is big
IF the generation is long AND the population
   size is medium THEN the crossover
   probability is very big
IF the generation is short AND the
   population size is big THEN the crossover
   probability is small
IF the generation is medium AND the
   population size is big THEN the crossover
   probability is medium
IF the generation is long AND the population
   size is big THEN the crossover
   probability is big
```
After the membership function and fuzzy rules have been defined, fuzzy rules can be developed to control crossover probability (Cp). The same way can be applied to control mutation probability (Mp) by using fuzzy rules as explained in the following table.

**Table 2. Fuzzy Decision of Mutation Probability (Mp)**

| Mutation Probability | Size Population | | |
|---|---|---|---|
| Generation | Small | Medium | Big |
| Short | Big | Medium | Small |
| Medium | Medium | Small | Very small |
| Long | Small | Very small | Very small |

Based on the data, it can be found 9 Fuzzy Rules as follows :
```
IF the generation is short AND the
   population size is small THEN the
   Mutation probability is big
IF the generation is medium AND the
   population size is small THEN the
   Mutation probability is medium
IF the generation is long AND the population
   size is small THEN the Mutation
   probability is small
IF the generation is short AND the
   population size is medium THEN the
   Mutation probability is medium
IF the generation is medium AND the
   population size is medium THEN the
   Mutation probability is small
IF the generation is long AND the population
   size is medium THEN the Mutation
   probability is small
IF the generation is short AND the
   population size is big THEN the Mutation
   probability is small
IF the generation is medium AND the
   population size is big THEN the Mutation
   probability is small
IF the generation is long AND the population
   size is big THEN the Mutation probability
   is small
```

Based on fuzzy inferential system with the Mamdani's method, the system will give a crossover probability output and a mutation. It is assumed in this research that the system is used to count the composition of food ingredients in order to determine the diet for the patients of kidney and urinary tract diseases. The diseases included in kidney and urinary tract diseases are nefrotic syndrome, acute renal failure, chronic kidney diseases, kidney transplantation and kidney failure with dialysis. There are some parameters used to determine the diet, such as weight, height, gender, age, activity factors and stress factors of a patient; all of which determines the amount of total energy required. The patient observed in this research was a male of 40 years old, 165cm/50kg. He was observed due to the kidney and urinary tract diseases. He was on a bed –rest with an activity factor of 1.2 and the stress factor of 1.4 (mild stress): the required energy for the AMB was calculated using formula (1). Meanwhile, the need of the total energy was used formula (3). Overall, the diet for the kidney disease focuses on controlling the consumption of energy, protein, liquid, and ions such as sodium, potassium, calcium, and phosphorous found in the daily food consumption. The food ingredients used consist of various categories; cereals, starch of various bulbs, nuts, vegetables, fruits, meat and poultry, fish, oyster, prawn, egg, milk, fat, vegetable oil, sugar, syrup, *koneksiori* and spices. In this research, 400 kinds of food from the survey were used. There was formerly a list of 600 kinds of food. The food database was created and could be accessed by calling the index. The chosen food ingredients were the combination of 10 best food ingredients that met the nutritional needs of the patient in a day. The food index served as a gene, and a chromosome consisted of 10 genes. For initializing the population, the food index was called randomly (score 1-400) in a 100 (for example) population or chromosomes; each chromosome consisted of 10 genes. In total, there were 1000 genes which were derived from randomizing the index. By doing this, it was possible that one or more indexes did not appear or appear more than once. For example, table 1 showed that the initial population had 20 chromosomes (shown with the number of rows. Each chromosome has 10genes. The number of the genes was shown by the number of the columns. Table 3 shows the initial population of 200 genes.

**Table 3. The initial population**

| No | 1 | 2 | 3 | 4 | 5 | 6 | 7 | 8 | 9 | 10 |
|---|---|---|---|---|---|---|---|---|---|---|
| 1 | 68 | 137 | 185 | 355 | 74 | 97 | 278 | 313 | 164 | 396 |
| 2 | 148 | 363 | 28 | 109 | 99 | 172 | 214 | 319 | 168 | 31 |
| 3 | 137 | 387 | 17 | 354 | 100 | 155 | 182 | 108 | 184 | 329 |
| 4 | 231 | 63 | 314 | 100 | 76 | 167 | 240 | 192 | 99 | 143 |
| 5 | 229 | 353 | 114 | 345 | 231 | 377 | 306 | 259 | 370 | 167 |
| 6 | 19 | 21 | 264 | 9 | 71 | 134 | 198 | 13 | 61 | 137 |
| 7 | 108 | 301 | 369 | 136 | 251 | 271 | 359 | 137 | 145 | 294 |
| 8 | 87 | 298 | 353 | 265 | 209 | 60 | 18 | 324 | 316 | 56 |
| 9 | 102 | 7 | 273 | 301 | 74 | 322 | 265 | 8 | 265 | 278 |
| 10 | 289 | 161 | 38 | 215 | 105 | 53 | 275 | 202 | 71 | 277 |
| 11 | 75 | 152 | 176 | 369 | 126 | 267 | 191 | 123 | 196 | 22 |
| 12 | 326 | 35 | 173 | 48 | 219 | 193 | 122 | 353 | 19 | 284 |
| 13 | 244 | 315 | 168 | 318 | 214 | 320 | 181 | 274 | 97 | 328 |
| 14 | 116 | 101 | 281 | 96 | 111 | 167 | 48 | 352 | 333 | 376 |
| 15 | 272 | 18 | 307 | 274 | 338 | 184 | 295 | 81 | 150 | 182 |
| 16 | 174 | 18 | 307 | 274 | 338 | 184 | 295 | 81 | 150 | 182 |
| 17 | 352 | 368 | 159 | 232 | 12 | 21 | 297 | 290 | 156 | 282 |
| 18 | 188 | 124 | 309 | 122 | 212 | 357 | 56 | 299 | 14 | 280 |
| 19 | 110 | 185 | 251 | 389 | 392 | 106 | 107 | 207 | 116 | 188 |
| 20 | 80 | 345 | 34 | 54 | 60 | 342 | 1 | 23 | 45 | 343 |

For instance, the fifth line is the fifth chromosome which has 10 gen consisting of the following food ingredient index; 229 (eggs derived from local chickens), 353 (condensed milk - unsweetened), 114 (soy milk), 345 (breast milk), 231 (Kentucky's fried chicken- breast), 377 (ginger), 306 (prawn), 259 (oyster), 370 (sweet chocolate bars) and 167 (red tomato). After the process of initialization, two chromosomes will be selected to be parents that will be crossly moved proportionally according to the fitness score. The selection method applied here is the roulette wheel selection method. The chromosomes having a higher fitness score will have a bigger chance compared to the chromosomes having a smaller fitness score.

A fitness score from a chromosome will show the quality of the chromosomes in the population. The diet's rule for each type of kidney and urinary tract diseases is various, so the guidance for the fitness score highly depends on the type of diets and the food composition. There are 5 types of diseases, i.e. : (1)patients with *nefrotic* syndrome are to take food rich in energy, protein, fat, and sodium; (2) patients with an acute kidney failure are to take food that contains energy, protein, and fat; (3)patients with chronic kidney diseases are to consume food which contains energy, protein, fat, and potassium; (4) patients at the final stage of kidney diseases are to consume food containing energy, protein, fat, calcium and phosphorous; (5) patients suffering from stone kidney are to consume food which has high energy, protein, carbohydrate, fat, calcium, and phosphorous. The example of data patterns which will be tested for the 5$^{th}$ chromosomes (see table 3) can be seen in table 4.

**Tabel 4. Sample of Data Test**

| No | Code | Materials | Composition of nutrient per 100 gram | | | | | | | |
|---|---|---|---|---|---|---|---|---|---|---|
| | | | Energy | Protein | Fat | Carbohydrate | Calcium | Phosphorous | Sodium | C Potassium |
| | | | kkal | G | G | G | mg | Mg | Mg | Mg |
| 229 | IDH001 | Eggs of local chicken | 174 | 11 | 14 | 1.2 | 68 | 268 | 190 | 141 |
| 353 | IDJ010 | Condensed milk- unsweetened | 138 | 7 | 7.9 | 9.9 | 243 | 195 | 140 | 303 |
| 114 | IDA005 | Soy milk | 41 | 3.5 | 2.5 | 5 | 50 | 45 | 0 | 0 |
| 231 | IDA057 | Fried chicken –Breast | 298 | 34 | 17 | 0.1 | 90 | 284 | 0 | 0 |

The fitness function used for *nefrotic* syndrome is:

$$f = \frac{1}{((abs(p-\sum a) + abs(q-\sum b) + abs(r-\sum c) + abs(s-\sum d)) + bilkecil)} \quad (4)$$

The fitness function used for acute renal failure is:

$$f = \frac{1}{((abs(p-\sum a) + abs(q-\sum b) + abs(r-\sum c)) + bilkecil)} \quad (5)$$

The fitness function used for chronic kidney diseases is:

$$f = \frac{1}{((abs(p-\sum a) + abs(q-\sum b) + abs(r-\sum c) + abs(t-\sum e)) + bilkecil)} \quad (6)$$

The fitness function used for the final stage of kidney diseases is:

$$f = \frac{1}{((abs(p-\sum a) + abs(q-\sum b) + abs(r-\sum c) + abs(v-\sum g) + abs(u-\sum f)) + bilkecil)} \quad (7)$$

The fitness function used for kidney stone is:

$$f = \frac{1}{((abs(p-\sum a)+abs(q-\sum b)+abs(w-\sum h)+abs(r-\sum c)+abs(v-\sum g)+abs(u-\sum f))+bilkecil)} \quad (8)$$

Explanation:
- p = The energy needed (calorie) or the whole energy is calculated based on operation 1-3.
- q = protein needed in 1 day based on the diet type
- r = fat needed in 1 day based on the diet type
- s = sodium needed in 1 day based on the diet type
- t = potassium needed in 1 day based on the diet type
- u = phosphorous needed in 1 day based on the diet type
- v = calcium needed in 1 day based on the diet type
- w = carbohydrate needed in 1 day based on the diet type

Based on the data in table 2, for eggs of local chicken:
- a = the calorie content is 174 kcal
- b = the protein content is 14 g
- d = the sodium content is 12 mg
- e = the potassium content is 68 mg
- f = the phosphorous content is 268 mg
- g = the calcium content is 190 mg
- h = the carbohydrate content is 141 g
- small of number = the number to avoid subtraction with zero

After going through a parental selection, then the crossover is conducted, with various scores of crossover probability. Crossover can only be operated with probability Pc. only if a derived random score [0 1] less than the defined Pc. An intersection is chosen randomly, and then the first part of parental 1 is combined with the second part of parental 2. After we obtain a crossover population, then the next is the process of mutation. In the process of mutation, the probability was needed to be set first. If the determined random score is less than the probability of mutation Pm, then change the genes from the food index that has not been appeared at the time of the initiation of population. Mutation will result in a new population, which will be the parents for the next generations. From the derived generations, the chromosomes which are potential to be the best chromosome are chosen. The best chromosome will be recommended as food index that will be consumed by patients suffering from kidney and urinary tract diseases in 1 day, because the food meets the calorie needs.

## 4.2 Test Results

The process of genetic fuzzy system comprises of using fuzzy logics for determining the precise crossover and mutation probability according to the rules. The inferential system applied is Mamdani's model. The input parameters the population is 100 and 150, meanwhile for the generation is 100, 800 and 1500. For the clearer explanation, see Table 5.

**Table 5. the parameters applied based on the Mamdani's fuzzy inferential system**

| The types of Diet | Population | Generation | Crossover Probability | Mutation Probability | Fitness Function |
|---|---|---|---|---|---|
| Nefrotik Syndrome | 100 | 100 | 0.168 | 0.489 | 0.000008 |
|  | 150 | 800 | 0.429 | 0.132 | 0.000817 |
|  | 150 | 1500 | 0.600 | 0.130 | **0.001666** |
| Acute Kidney Failure | 100 | 100 | 0.168 | 0.489 | 0.002018 |
|  | 150 | 800 | 0.429 | 0.132 | 0.001702 |
|  | 150 | 1500 | 0.600 | 0.130 | **0.002168** |
| Chronic Kidney Diseases | 100 | 100 | 0.168 | 0.489 | **0.001476** |
|  | 150 | 800 | 0.429 | 0.132 | 0.001281 |
|  | 150 | 1500 | 0.600 | 0.130 | 0.000988 |
| The Final Stage of Kidney Diseases | 100 | 100 | 0.168 | 0.489 | 0.000570 |
|  | 150 | 800 | 0.429 | 0.132 | **0.001221** |
|  | 150 | 1500 | 0.600 | 0.130 | 0.001144 |
| Kidney Stone Diseases | 100 | 100 | 0.168 | 0.489 | 0.000007 |
|  | 150 | 800 | 0.429 | 0.132 | 0.000552 |
|  | 150 | 1500 | 0.600 | 0.130 | **0.000587** |

As shown in the table, the best fitness function is printed in bold. From the result of the three types of tests to each kind of diet, a parameter is chosen. The chosen parameter may result in the best fitness function. The test for nefrotic syndrome diet uses the population size of 150, crossover probability of 0.6, mutation probability of 0.13 and generation maximum of 1500. The food ingredient index -resulted from the trial error test using the parameter- is 372, 327, 72, 398, 196, 386, 344, 195, 109, 134; and with the fitness function of 0.006978. The food ingredient composition is as follows: palm sugar (13 gram), anchovy *balado* –spicy anchovy- (25 gram), cassava *getuk –cassava cake with shredded coconut*- (100 gram), soya sauce (not included in the rule since it is a non energy food), Mango *harum manis* –fragrant and sweet mango-(90 gram), fresh tamarind, skimmed milk powder 30 gram) , mango (90 gram), *oncom* (25 gram), red spinach (100 gram). The following is the graph showing the test result.

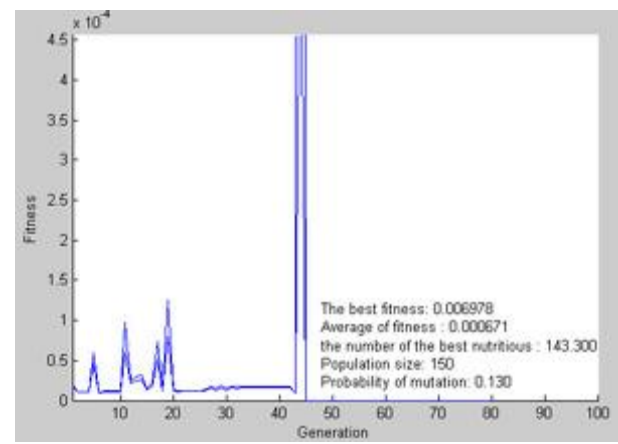

**Fig 7 : The test result for nefrotic syndrome diet**

The test for acute kidney failure uses the population size of 150, crossover probability of 0.6, mutation probability of 0.13 and maximum generation of 1500. The index of food ingredients resulted from the test using the parameter is 59, 109, 95, 147, 312, 359   219, 164, 255, 231 with the fitness function of 0.000920. The compositions of the food ingredients consist of steamed yellow sweet potato (135 gram), *oncom* (25 gram), boiled nuts (20 gram), young corns (100 gram), *gulai* oyster –oyster in spicy coconut sauce- (60 gram), fats of cow (50 gram), duck meat (45 gram), boiled lettuce (because it contains very small amount of energy, protein, and carbohydrate so lettuce can be consumed anytime, snail (50 gram), Kentucky fried chicken –drum stick (50 gram). The graph showing the result of the test can be seen in the Fig.8.

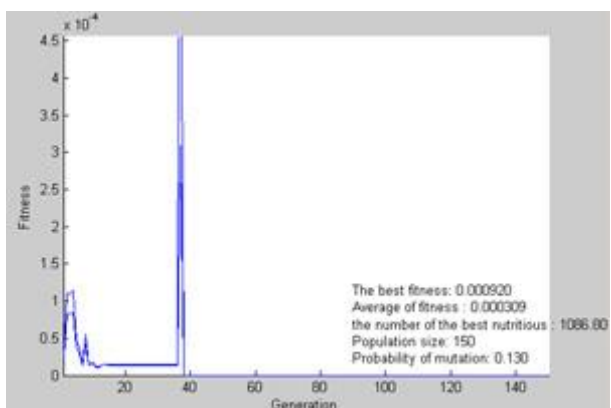

**Fig 8 : The result of the test for the kidney failure diet**

The test for chronic kidney diet is used for the population size of 0.168, mutation probability of 0.489 and maximum generation of 100. The index resulted from the test using the parameter is 324, 143, 13, 98, 214   268, 364, 305, 389, 5 with the fitness function of 0.000999. The food ingredients consist of salted cashew nut (40 gram), papaya leaves (100 gram), rice cracker, young coconut flesh (30 gram), young *sukun* fruit –jack fruit-, fried *mujahir* fish –tilapia (50 gram), soy bean oil (5 gram), dried shrimp (50 gram), onion, black glutinous rice (400 gram). The graph showing the result is as in Fig.9.

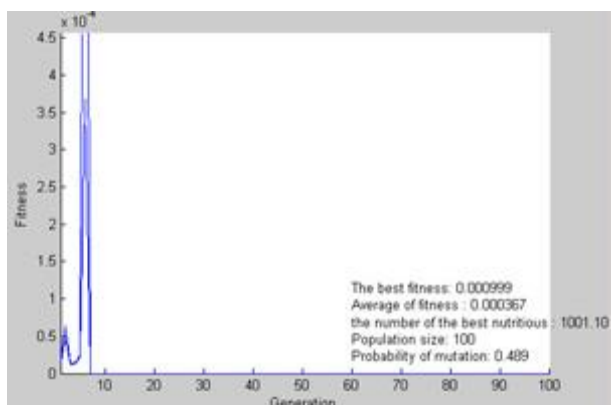

**Fig 9 : The result of the test using cronic kidney diseases diet**

The test for the final phase of kidney diet uses the population size of 150, crossover probability of   0.429, mutation probability of 0.139 and maximum generation of 800. The index resulted from the test using the parameter is 71, 344, 171, 17, 162, 111, 142, 146, 352, 85; with the fitness function of 0.000488. The composition of the food ingredients is fried *getuk* (100 gram) , milk (100 gram), steamed carrot (100 gram), macaroni (50 gram), seaweed, pure coconut milk, cabbage (100 gram), boiled cassava leaves (100 gram), sweetened condensed milk 100 gram), black beans (10 gram). The result of the test is shown in the Fig.10.

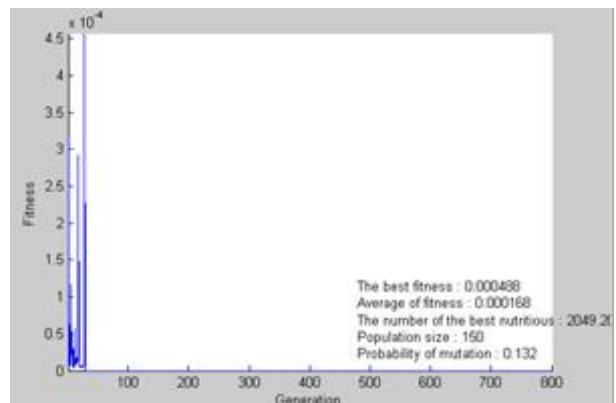

**Fig 10 :** The result of the test for the kidney diet in the final phase

The test for the kidney stone uses the population size of 150, crossover probability of 0.6, mutation probability of 0.13 and maximum generation of 1500. The index resulted from the test using the parameter is 55, 356, 187, 371, 1, 363, 138, 88, 166, and 261 with the fitness function of 0.000692. The followings are the food ingredients: potato (200 gram), yoghurt (100 gram), Bali orange (100 gram), chocolate milk bar, ground rice (100 gram), nuts oil (5 gram), papaya flowers (100 gram), boiled soya beans (100 gram), bean sprouts (100 gram), *layur* fish –eel fish (15 gram). The graph is shown in Fig.11.

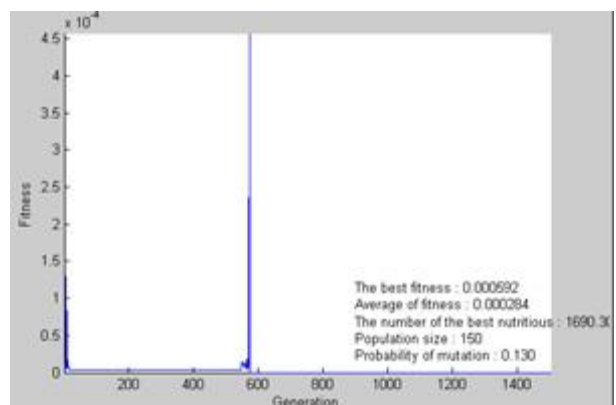

**Fig 11 ; The result of the test for kidney stone diet**

The population size used for the five types of above diets resulting in the best fitness function is 150 except for the chronic kidney diet. However, the amount of the generation used greatly varied.

## 5.   CONCLUSION

The real test cases shows that genetic fuzzy algorithm can be applied to determine the composition of the optimal food ingredients to fulfill the nutrition need in a day for patients with kidney and urinary tract diseases. The kinds of diseases observed in this research are the nefrotic syndrome, the acute kidney failure, the chronic kidney, the final phase of kidney disease and kidney stone. The best fitness function means the composition from the 10 food ingredients resulting in the nutrition value to fulfill the minimum diet requirement to the patients with kidney and urine tract diseases. However, further investigations on extending the disease database by incorporating various other types of diseases, developing friendly user interface, and improving the method by combining other techniques of soft computing are required disease, build a user-friendly user interface and improving the methods by combining the techniques of other soft computing.